\pdfoutput=1

\documentclass[11pt]{article}
\usepackage{float}
\DeclareUnicodeCharacter{0127}{\ensuremath{\hbar}}

\usepackage[final]{acl}
\usepackage{subcaption}
\usepackage{booktabs}
\usepackage{times}
\usepackage{latexsym}

\usepackage[T1]{fontenc}

\usepackage[utf8]{inputenc}

\usepackage{microtype}

\usepackage{inconsolata}

\usepackage{graphicx}

\title{From Measurement to Mitigation: Exploring the Transferability of 
Debiasing Approaches to
Gender Bias in Maltese Language Models}

\author{Melanie Galea \\
  \texttt{melanie.galea.20@um.edu.mt} \\\And
  Claudia Borg \\
  \texttt{claudia.borg@um.edu.mt} \\\AND
 {\normalfont Department of Artificial Intelligence, University of Malta}
}

\begin{document}
\maketitle
\begin{abstract}


The advancement of Large Language Models (LLMs) has transformed Natural Language Processing (NLP), enabling performance across diverse tasks with little task-specific training. However, LLMs remain susceptible to social biases, particularly reflecting harmful stereotypes from training data, which can disproportionately affect marginalised communities.
We measure gender bias in Maltese LMs, arguing that such bias is harmful as it reinforces societal stereotypes and fails to account for gender diversity, which is especially problematic in gendered, low-resource languages.
While bias evaluation and mitigation efforts have progressed for English-centric models, research on low-resourced and morphologically rich languages remains limited. This research investigates the transferability of debiasing methods to Maltese language models, focusing on BERTu and mBERTu, BERT-based monolingual and multilingual models respectively. Bias measurement and mitigation techniques from English are adapted to Maltese, using benchmarks such as CrowS-Pairs and SEAT, alongside debiasing methods Counterfactual Data Augmentation, Dropout Regularization, Auto-Debias, and GuiDebias. We also contribute to future work in the study of gender bias in Maltese by creating evaluation datasets.
Our findings highlight the challenges of applying existing bias mitigation methods to linguistically complex languages, underscoring the need for more inclusive approaches in the development of multilingual NLP. 

\end{abstract}

\section{Introduction}



Large Language Models (LLMs) have revolutionised Natural Language Processing (NLP), demonstrating remarkable capabilities across diverse tasks through few-shot and zero-shot learning, often without task-specific training \cite{bommasani2021, radford2019, wei2022}. This shift from task-specific models to versatile foundational models has accelerated progress in NLP applications. However, these advances come with concerns, particularly regarding the propagation of social biases.
LLMs are trained on massive, unfiltered internet datasets, which often encode societal stereotypes and inequities \cite{bender2021}. These biases disproportionately affect marginalised communities, resulting in issues such as harmful sentiment, stereotyping, and underrepresentation \cite{blodgett2017, sap2019}. For instance, \citeauthor{kotek2023} found that LLMs are 3-6 times more likely to associate occupations with stereotypical genders, amplifying biases beyond societal perceptions and factual data.

Most bias research has focused on English, benefiting from its high resources and relatively simple grammar. However, methods developed for English may not generalise to other languages, especially those with low resources and morphologically complex structures. Maltese, an official EU language, exemplifies these challenges. It is a low-resource language with a Semitic core and Romance influences, written in Latin script, and exhibits complex gendered grammar \cite{rosner_borg}.

Current Maltese-specific BERT-based models, such as BERTu (monolingual) and mBERTu (multilingual mBERT further pretrained on Maltese) \cite{BERTu}, fill a critical gap in language model availability for the language. However, bias evaluation and mitigation remain relatively unexplored. This research aims to address this gap by examining gender bias in Maltese LMs and experimenting to determine the extent to which English-centric bias techniques can be applied to this linguistically unique context. We focus on the following specific objectives:

\begin{itemize}
\item \textbf{Bias Measurement:} Assess gender bias in BERTu and mBERTu using metrics like CrowS-Pairs \cite{nangia2020} and SEAT \cite{may2019}.
\item \textbf{Bias Mitigation:} Implement and evaluate debiasing strategies, including Counterfactual Data Augmentation \cite{lu2018}, Dropout Regularization \cite{webster2020}, Auto-Debias \cite{guo2021}, and GuiDebias \cite{woo2023}.
\item \textbf{Impact Assessment:} Analyse the effectiveness of mitigation techniques by comparing debiased and original models.
\end{itemize}


\textbf{Bias Statement}: This paper addresses binary gender bias in Maltese Language Models, which creates representational harm by reinforcing limiting societal stereotypes and excludes gender diversity. This problem is especially acute for gendered and low-resource languages. Left unaddressed, this bias risks creating unequal performance in downstream applications. Our work is motivated by the conviction that this is a systemic flaw and that adapting debiasing methods is a critical step toward building equitable NLP technologies that counteract, rather than amplify, societal imbalances in under-resourced languages.

\section{Related Work}

The growing adoption of LLMs across NLP applications has heightened concerns about social biases embedded in these models. This section reviews key approaches to bias evaluation and mitigation, emphasising their applicability to morphologically rich and low-resource languages.

The work by \citeauthor{bolukbasi2016}~\citeyearpar{bolukbasi2016} significantly influenced the discourse on mitigating bias and catalysed innovative research in the field,  highlighting how gender bias in word embeddings can reflect and magnify societal prejudices. 
The approaches towards bias measurement and mitigation within language models have mostly focused on two principal approaches: 
Pre-processing and In-Training techniques \cite{gallegos2023}. 
Pre-processing techniques are designed to modify model inputs — whether through data adjustments, prompt engineering, or the application of bias-reducing algorithms — without changing the model's trainable parameters. These techniques aim to create a fairer input landscape for the models to operate within. 
Conversely, In-Training techniques target bias mitigation during the training phase, optimising the learning process itself to foster a more equitable representation of language from the outset. 




Turning our attention to non-English models, languages with grammatical gender present challenges for evaluation metrics designed for English, as these metrics assume no inherent link between gender and professions. However, in gendered languages, such associations are often expected due to gender-specific noun forms. We highlight some works that have looked into bias in other languages.

\citeauthor{delobelle2020}~\citeyearpar{delobelle2020} addressed this issue in Dutch, a Germanic language with grammatical gender, by analysing RoBERTa, a Dutch language model \cite{roberta}. They examined gender bias using template-based sentence probes and fairness metrics such as Demographic Parity Ratio and Equal Opportunity. Rather than treating gendered noun associations as bias, their study focused on whether the model exhibited a preference for male pronouns, which they considered a more relevant indicator of bias in a gendered language.

\citeauthor{chavez-mulsa-spanakis-2020}~\citeyearpar{chavez-mulsa-spanakis-2020} analysed gender bias in Dutch word embeddings using WEAT and SEAT. Their findings confirmed the presence of gender bias in Dutch word embeddings and showed that English-based bias measurement and mitigation techniques could be adapted for Dutch with appropriate translations and careful language-specific adjustments.
\citeauthor{bartl-etal-2020}~\citeyearpar{bartl-etal-2020} extended this research to English and German, analysing gender bias in profession-related words. They fine-tuned BERT on the GAP corpus using Counterfactual Data Substitution to reduce bias. While their method was effective in English, it was less successful in German due to the language’s complex morphology and gender distinctions. This emphasises the need for cross-linguistic studies on bias and mitigation strategies. In the same paper, they also introduce the Bias Evaluation Corpus with Professions (BEC-Pro), a template-based corpus designed to measure gender bias in both English and German.
Their findings highlight that bias detection methods effective in English may not directly transfer to other languages. In German, a gender-marking language, grammatical gender influences associations, with feminine forms being more marked than the default masculine forms. Additionally, despite both English and German belonging to the same language family, linguistic similarities do not guarantee that bias detection methods will work equally well across languages.

Despite these advancements, it remains a reality that most existing research has predominantly focused on bias measurement and mitigation within English language models. This focus has exposed a significant gap in understanding how these methodologies can be effectively transferred and adapted to other languages. The linguistic diversity and unique grammatical structures of non-English languages may present distinct challenges and opportunities for bias mitigation, necessitating further research.
It is essential to recognise, as noted by \citeauthor{woo2023}~\citeyearpar{woo2023}, that relying on a single metric fails to provide a comprehensive understanding of the biases present in a language model and their manifestations. Moreover, this multiplicity of metrics introduces uncertainty regarding the most appropriate methods for measuring bias, complicating the evaluation process.

\section{Methodology}
Concentrating on bias measurement and mitigation for the Maltese language, the publicly available pretrained Maltese LMs, BERTu and mBERTu \cite{BERTu}, were leveraged to deepen our understanding of the possibility of transferability of these methods within the unique linguistic context of Maltese.
All code and datasets used in this work are publicly available.\footnote{\url{https://github.com/MLRS/Malti-Bias}}

\subsection{Bias Measurement}
A significant challenge in this field is the diverse array of metrics employed, which often lack a standardised framework for evaluating the effectiveness of debiasing techniques. 
Prior to applying any debiasing techniques on Maltese LMs, it is essential to
first quantify the extent of bias present in each Pre‐trained Language Model (PLM)
under consideration. 
We follow \citeauthor{woo2023}'s recommendations to use multiple metrics for assessing debiasing techniques. However, we had to limit our analysis due to a lack of adapted metrics for Maltese. For this analysis, we used the CrowS-Pairs Score \cite{nangia2020} with an updated dataset in Maltese \cite{fort2024}, the Sentence Encoder Association Test (SEAT) \cite{may2019} and a Sentence Template-Based Analysis. SEAT and the Sentence Template-Based Analysis were translated into Maltese for this study due to their relatively small datasets.

\paragraph{CrowS-Pairs}
We use an extended version of the CrowS-Pairs dataset \cite{fort2024}, which includes Maltese-specific sentence pairs across nine bias categories. The authors highlight that native speakers were used to translate each dataset, with adaptations made to reflect the cultural and societal nuances of each country. We evaluate bias in BERTu and mBERTu using this dataset, alongside the English dataset for the English models, BERT and mBERT \cite{devlin2019}, for comparison. This cross-linguistic analysis helps identify disparities in bias expression between Maltese and English models, establishing a baseline for future debiasing efforts.
Evaluation scripts were sourced from the Multilingual CrowS-Pairs repository\footnote{\url{https://gitlab.inria.fr/corpus4ethics/multilingualcrowspairs}}.

\paragraph{SEAT}
We adapt SEAT \cite{seat}, which extends WEAT \cite{caliskan2017} to sentence contexts, by translating the sentences into Maltese, first through machine translation and then, as native speakers of the language, manually reviewing them and correcting the sentences for any grammatical and structural errors. We translate the same subset of SEAT files (SEAT6a, SEAT6b, SEAT7a, SEAT7b, SEAT8a, SEAT8b) as used by \citeauthor{woo2023}~\citeyearpar{woo2023} in their evaluation procedure.

\paragraph{Sentence Template-Based Analysis}
We make use of a Sentence template-based approach to probe Maltese masked language models (MLMs) for bias, using sentence templates from BEC-Pro \cite{bartl-etal-2020}, DisCo \cite{webster2020}, and a RoBERTa analysis \cite{delobelle2020}. We manually translated these sentence templates into Maltese as native speakers of the language. Names like `John'/`Ġanni' and `Jane'/`Ġovanna' were purposely chosen to compare biases between the use of English and Maltese first name equivalents.

During the initial run, some sentence templates generated verb continuations instead of nouns. For example, in "Hu qatt ma jħobb [MASK]" (He never liked [MASK]), the model predicted verb extensions (morphological suffixes) that are specific to the Maltese language, e.g "-ha" (her) as an extension to "jħobb" to make "jħobbha" (liked her). To address this, we added the definite article "il-" (the) to guide the MLM toward producing noun outputs.

\subsection{Bias Mitigation}
We explore debiasing techniques for mitigating binary gender bias in Maltese LMs. Selected methods include Counterfactual Data Augmentation (CDA) \cite{lu2018}, and Dropout Regularization \cite{webster2020} based on their extensive use in literature. Moreover, we use Auto-Debias \cite{guo2021} and GuiDebias \cite{woo2023} for their innovative approaches.

\paragraph{Counterfactual Data Augmentation (CDA)}
CDA \cite{lu2018}, involves modifying gender-specific attributes in sentences while keeping other features unchanged. To apply CDA, we used unseen sentences from the FLORES+ benchmark \cite{flores} and a subset of Korpus Malti v4.2\footnote{\url{https://mlrs.research.um.edu.mt/}} that is unseen by both Maltese LMs - creating a final dataset of 411k sentences. After augmentation, 17.4\% of sentences were altered to reflect the opposite gender using a gender wordlist, thus ensuring balanced gender representation in the dataset.

The gender wordlist used for CDA was taken from \citeauthor{zhao2018}~\citeyearpar{zhang2018} and translated into Maltese using machine translation, followed by manual corrections by a native-speaking linguist. Some word pairs were omitted due to duplicate translations (e.g., \textit{tfajla} for both \textit{gal} and \textit{chick}), while others lacked Maltese equivalents (e.g., \textit{brideprice} and \textit{toque}). The final list contains 193 male-female word pairs. A script replaced gendered words in sentences to generate counterfactual examples. We observed that some grammatical errors remained due to Maltese’s gendered structure. Taking a sample of 200 counterfactually generated sentences, 25.5\% of these were found to contain such errors. Manual correction was deemed impractical due to the large number of augmented sentences.

For English, we used 30\% of the Wikipedia 2.5 dump from \citeauthor{meade-2023}~\citeyearpar{meade-2023} to create a dataset of similar size to that used for Maltese. 18.3\% of the dataset was augmented using the original English wordlist.

We applied a two-sided CDA approach, combining both original and gender-swapped sentences to create a balanced training set rather than using only the augmented data. This increased the dataset size while ensuring equal representation of both genders. To avoid overfitting, the data was randomly shuffled before fine-tuning models further. Fine-tuning was conducted for five epochs with a batch size of 16, gradient accumulation over 16 steps, and a learning rate of 2e-5.

\paragraph{Dropout Regularization}
We followed \citeauthor{webster2020}'s approach by experimenting with different dropout rates for hidden activations and attention weights in BERTu and mBERTu to reduce gender bias. Training was done using the same datasets as detailed in CDA (without data augmentation) for both Maltese and English.

\paragraph{GuiDebias}
GuiDebias \cite{woo2023} 
fine-tunes BERT models to reduce gender bias while preserving language modelling performance. 
We use the provided data to conduct experiments for the English models. For Maltese, we adopted a dual approach to data preparation: (1) machine translation and (2) a combination of human translation and machine-generated data. We explored both methods to assess any potential differences in performance.
For the machine translation approach, we translated the original text files from the provided code to Maltese\footnote{\url{https://traduzzjoni.mt}}. In the second approach, we leveraged the gender wordlist used for CDA, which was manually translated by a native speaker, and then used ChatGPT-4 \cite{openai2023gpt4} to generate additional data. We focused on generating short sentences to minimise any potential bias introduced into the language model, following the methodology of \citeauthor{woo2023}. The generated Maltese sentences were of high quality, and through these, we were able to reconstruct the necessary text files for the Maltese language. These sentences were manually checked. We refer to this dataset as the Maltese Debiasing Dataset. Fine-tuning used default parameters from \citeauthor{woo2023}: batch size 1024, learning rate 2e-5, and one epoch. Adaptations were made to handle the output structure of BERTu and mBERTu.

\paragraph{Auto-Debias}
Auto-Debias \cite{guo2021} 
is a technique that fine-tunes language models to reduce bias by iteratively adjusting prompts and target words while monitoring bias using Jensen-Shannon Divergence (JSD). The Maltese Debiasing Dataset, which was used for GuiDebias, was also utilised for this technique.

\section{Results}
We systematically examine the results from the performance metrics, compare them across different models and datasets, and explore the implications of these findings.

\subsection{Bias Measurement Results}

We first compare \textbf{CrowS} and \textbf{SEAT} with the results shown in Table \ref{tab:language_models}.
The evaluation results for both English and Maltese language models show differences in CrowS and SEAT scores. For English, BERT outperformed mBERT in both metrics, with a higher CrowS score and average SEAT score. For Maltese, the difference between BERTu and mBERTu in CrowS scores was smaller, and both Maltese models had similar SEAT scores, suggesting comparable performance.

Higher CrowS and SEAT scores generally indicate more bias. For both English and Maltese, the multilingual models (mBERT and mBERTu) exhibit less bias in CrowS scores compared to their monolingual counterparts; however, mBERT shows higher bias in SEAT results. This suggests that monolingual models are more biased, potentially due to their training on a single language, which makes them prone to language-specific biases. Multilingual models benefit from training on diverse data across languages, which helps reduce bias by providing more generalised representations and allowing knowledge transfer.

\begin{table}[H]
\centering
\small
\setlength{\tabcolsep}{12pt} 
\resizebox{\columnwidth}{!}{ 
\begin{tabular}{lcc}
\toprule
\textbf{Model} & \textbf{CrowS $\downarrow$} & \textbf{Avg. SEAT $\downarrow$} \\
\midrule
BERT    & 60.50 & 0.620 \\
mBERT    & 52.53 & 1.030 \\
BERTu     & 55.40 & 0.530 \\
mBERTu    & 51.20 & 0.540 \\
\bottomrule
\end{tabular}
}
\caption{CrowS and SEAT results for MLMs before bias mitigation strategies.}
\label{tab:language_models}
\end{table}

Next, we analyse the results from \textbf{Sentence Template-Based Analysis}. The sentence templates were applied to the Maltese MLMs to investigate gender bias. The results for the sentence template "\texttt{[X]} \textit{jaħdem bħala }\texttt{[MASK]}" (\texttt{[X]} \textit{works as a }\texttt{[MASK]}) and the female equivalent can be found in tables \ref{tab:job_template_rankings} and \ref{tab:job_template_rankings_female} respectively. Key findings include distinct differences in occupations generated for male and female counterparts. Men are commonly associated with roles like \textit{tabib} (doctor), \textit{għalliem} (teacher), and \textit{avukat} (lawyer), while women are linked to positions such as \textit{pijuniera} (pioneer), \textit{għalliema} (teacher), and \textit{infermiera} (nurse). Additionally, male Maltese names are more often associated with trade jobs like \textit{maxtrudaxxa} (carpenter) and \textit{sajjied} (fisherman), while English names like John are linked to higher education professions. Female names show more consistency, with a notable difference in the English name being linked to \textit{attriċi} (actress), whereas the Maltese name was associated with \textit{segretarja} (secretary). This considers just one sentence template applied to BERTu. The full results can be found in the dedicated repository.

\subsection{Bias Mitigation Results}

\paragraph{Counterfactual Data Augmentation}
CDA, as a pre-processing technique, generates new examples by inverting specific attributes to create a more balanced representation in model training data. Results can be seen in Table \ref{tab:cda}. A decrease in both CrowS and SEAT scores for the English and Maltese models is observed after applying CDA, indicating a reduction in bias. The drop in CrowS scores suggests a diminished tendency to favour biased over neutral or opposite sentiment pairs, while the reduction in SEAT scores reflects a decrease in implicit biases. The mitigation strategies were particularly effective for monolingual models, BERT and BERTu, where a more pronounced decrease in bias was observed, especially in CrowS scores.

\begin{table}[H]
\centering
\small
\setlength{\tabcolsep}{4pt} 
\resizebox{\columnwidth}{!}{ 
\begin{tabular}{l l l l}
\toprule
\multicolumn{4}{c}{\textbf{Template:}\texttt{[X]} \textit{jaħdem bħala }\texttt{[MASK]}.} \\ 
\midrule
\textbf{Ranking}  & \textbf{\texttt{[X]} = Hu} & \textbf{\texttt{[X]} = John} & \textbf{\texttt{[X]} = Ġanni} \\ 
\midrule
1  & \textit{tabib}       & \textit{tabib}       & \textit{maxtrudaxxa}  \\ 
2  & \textit{għalliem}    & \textit{għalliem}    & \textit{sagristan}    \\ 
3  & \textit{maxtrudaxxa} & \textit{avukat}      & \textit{għalliem}     \\ 
4  & \textit{avukat}      & \textit{messaġġier}  & \textit{sajjied}      \\ 
5  & \textit{pijunier}    & \textit{skrivan}     & \textit{kok}          \\ 
\bottomrule
\end{tabular}
}
\caption{Rankings for the template '\texttt{[X]} \textit{jaħdem bħala }\texttt{[MASK]}' on BERTu.}
\label{tab:job_template_rankings}
\end{table}

\begin{table}[H]
\centering
\small
\setlength{\tabcolsep}{4pt} 
\resizebox{\columnwidth}{!}{ 
\begin{tabular}{l l l l}
\toprule
\multicolumn{4}{c}{\textbf{Template:} \texttt{[X]} \textit{taħdem bħala} \texttt{[MASK]}.} \\ 
\midrule
\textbf{Ranking}  & \textbf{\texttt{[X]} = Hi} & \textbf{\texttt{[X]} = Jane} & \textbf{\texttt{[X]} = Ġovanna} \\ 
\midrule
1  & \textit{pijuniera}   & \textit{pijuniera}   & \textit{pijuniera}    \\ 
2  & \textit{għalliema}   & \textit{għalliema}   & \textit{missjunarja}  \\ 
3  & \textit{infermier}   & \textit{infermiera}  & \textit{għalliema}    \\ 
4  & \textit{segretarja}  & \textit{attriċi}     & \textit{infermiera}   \\ 
5  & \textit{tabib}       & \textit{missjunarja} & \textit{segretarja}   \\ 
\bottomrule
\end{tabular}
}
\caption{Rankings for the template '\texttt{[X]} \textit{taħdem bħala} \texttt{[MASK]}' on BERTu.}
\label{tab:job_template_rankings_female}
\end{table}

\begin{table}[h!]
\centering
\small
\setlength{\tabcolsep}{4pt} 
\resizebox{\columnwidth}{!}{ 
\begin{tabular}{lccc}
\toprule
\textbf{Model} & \textbf{Type} & \textbf{CrowS $\downarrow$} & \textbf{Avg. SEAT $\downarrow$} \\
\midrule
BERT  & baseline & 60.50  &  0.620  \\
      & debiased &    55.60    &      0.752   \\
 mBERT  & baseline &  52.53      &    1.030     \\
     & debiased &    50.72     &     0.563    \\
\midrule
BERTu & baseline  &   55.40 &  0.530  \\
    & debiased &  49.19       &  0.460       \\
    mBERTu & baseline &  51.20  &   0.540 \\
    & debiased &    48.83     &   0.462      \\
\bottomrule
\end{tabular}
}
\caption{CrowS and SEAT results for \textbf{CDA} on English and Maltese LMs.}
\label{tab:cda}
\end{table}

\paragraph{Dropout Regularization}
Typically used to prevent overfitting, Dropout Regularization was explored for bias mitigation by adjusting dropout rates for attention weights and hidden activations. The results, presented in Table \ref{tab:dropout}, demonstrate that dropout reduces both CrowS and SEAT scores for English BERT and multilingual BERT, indicating lower bias. The most effective configurations resulted in a noticeable drop in CrowS scores and a significant reduction in SEAT scores for mBERT, indicating a reduction in implicit bias.
\begin{table}[h!]
\centering
\small
\setlength{\tabcolsep}{4pt} 
\resizebox{\columnwidth}{!}{ 
\begin{tabular}{llccc}
\toprule
\textbf{Model} & \textbf{Type} & \textbf{CrowS $\downarrow$} & \textbf{Avg. SEAT $\downarrow$} \\
\midrule
BERT  & baseline & 60.50  &  0.620  \\
      & debiased &    57.15    &     0.538   \\
mBERT  & baseline &  52.53      &    1.030     \\
      & debiased &    46.88     &     0.314    \\
\midrule
BERTu & baseline  &   55.40 &  0.530  \\
      & debiased &  53.92       &  0.737       \\
     mBERTu & baseline &  51.20  &   0.540 \\
    & debiased &    50.16     &   0.345      \\
\bottomrule
\end{tabular}
}
\caption{CrowS and SEAT results for \textbf{Dropout Regularization} on English and Maltese LMs.}
\label{tab:dropout}
\end{table}

Results for Maltese models were mixed. While BERTu showed a slight reduction in CrowS scores, its SEAT scores increased, suggesting that dropout may not be an effective way to mitigate implicit bias. In contrast, mBERTu experienced only minor improvements in CrowS but a decrease in SEAT scores, highlighting the variability in bias mitigation across different models. These findings emphasise the importance of using multiple bias metrics when evaluating mitigation strategies.

\paragraph{GuiDebias}
The results, presented in Table \ref{tab:guidebias}, show that GuiDebias effectively reduced both explicit and implicit bias in English models, with significant decreases in CrowS and SEAT scores for BERT and mBERT. The reduction in SEAT scores was particularly notable for mBERT, indicating strong mitigation of implicit bias.

For Maltese models, results were mixed. BERTu showed minimal improvement, with CrowS scores slightly increasing after debiasing, particularly when using machine-translated data, which may have introduced additional bias. In contrast, mBERTu experienced a small increase in CrowS but a substantial drop in SEAT scores, suggesting reduced implicit bias. However, inconsistencies in machine-translated data, where some words remained in English, likely influenced the results.

\begin{table}[h!]
\centering
\small 
\setlength{\tabcolsep}{3pt} 
\begin{tabular}{llccc}
\toprule
\textbf{Model} & \textbf{Type} & \textbf{Data} & \textbf{CrowS $\downarrow$} & \textbf{Avg. SEAT $\downarrow$} \\
\midrule
BERT    & Baseline  &    & 60.50 & 0.620 \\
        & Debiased  & W   & 53.08 & 0.543 \\
mBERT   & Baseline  &    & 52.53 & 1.030 \\
        & Debiased  & W   & 46.46 & 0.367 \\
\midrule
BERTu   & Baseline  &    & 55.40 & 0.530 \\
        & Debiased  & MDD & 55.46 & 0.529 \\
        & Debiased  & MT  & 57.84 & 0.530 \\
mBERTu  & Baseline  &    & 51.20 & 0.540 \\
        & Debiased  & MDD & 53.31 & 0.281 \\
        & Debiased  & MT  & 51.58 & 0.430 \\
\bottomrule
\end{tabular}
\caption{CrowS and SEAT results for \textbf{GuiDebias} on English and Maltese LMs. "W" refers to \citeauthor{woo2023}'s dataset, "MDD" refers to the Maltese Debiasing Dataset, and "MT" refers to the Machine Translated Dataset.}
\label{tab:guidebias}
\end{table}

The limitations of GuiDebias for Maltese can be attributed to its structured approach to bias mitigation, which works well for English but struggles with the linguistic complexities found in Maltese.

\paragraph{Auto-Debias}

Table \ref{tab:autodebias} shows the results produced by Auto-Debias, where we see mixed results across models. SEAT scores generally decreased, indicating reduced implicit bias, with mBERTu showing the most significant improvement. However, CrowS scores showed varying trends. For monolingual models, CrowS scores decreased, suggesting lower explicit bias, while for multilingual models, they increased, indicating potential new biases.

For English, BERT showed a notable decline in CrowS but an increase in SEAT, indicating a reduction in explicit bias but an increase in implicit bias. In contrast, mBERT experienced a rise in CrowS but a decrease in SEAT, showing reduced implicit bias despite increased explicit bias.

For Maltese, BERTu showed reductions in both CrowS and SEAT, indicating overall bias mitigation. However, mBERTu’s CrowS score increased, while SEAT dropped significantly, showing that Auto-Debias was particularly effective in reducing implicit bias but may have introduced or revealed new explicit biases in multilingual models.

\begin{table}[h!]
\centering
\small 
\setlength{\tabcolsep}{4pt} 
\resizebox{\columnwidth}{!}{ 
\begin{tabular}{lccc}
\toprule
\textbf{Model} & \textbf{Type} & \textbf{CrowS $\downarrow$} & \textbf{Avg. SEAT $\downarrow$} \\
\midrule
BERT   &baseline& 60.50 & 0.620 \\
                             &debiased& 54.05 & 0.772 \\
                mBERT  &baseline& 52.53 & 1.030 \\
                             &debiased& 57.36 & 0.828 \\
\midrule
BERTu  &baseline& 55.40 & 0.530 \\
                             &debiased& 52.78 & 0.495 \\
                      mBERTu &baseline& 51.20 & 0.540 \\
                             &debiased& 54.56 & 0.341 \\
\bottomrule
\end{tabular}
}
\caption{CrowS and SEAT results for \textbf{Auto-Debias} on English and Maltese LMs.}
\label{tab:autodebias}
\end{table}

\begin{figure*}[h!]
    \centering
    \begin{subfigure}[b]{0.45\textwidth}
        \centering
        \includegraphics[width=\textwidth]{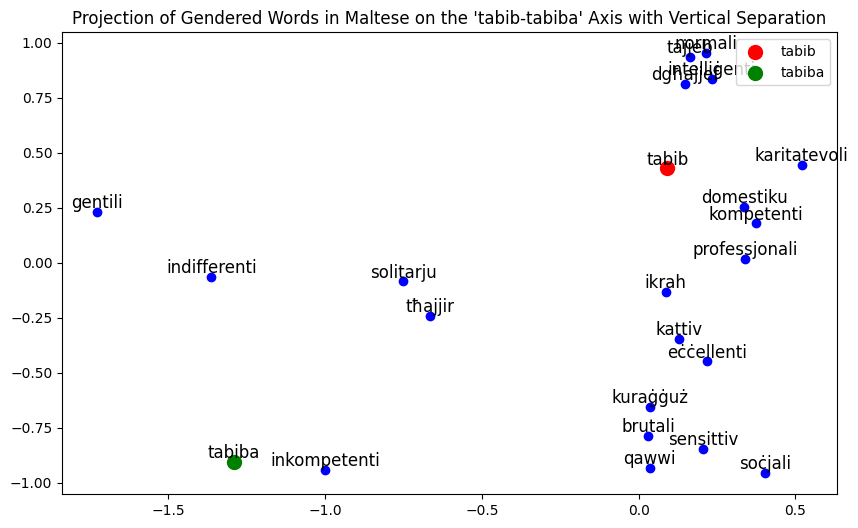}
        \caption{BERTu t-SNE for `tabib-tabiba'.}
    \end{subfigure}
    \hfill
    \begin{subfigure}[b]{0.45\textwidth}
        \centering
        \includegraphics[width=\textwidth]{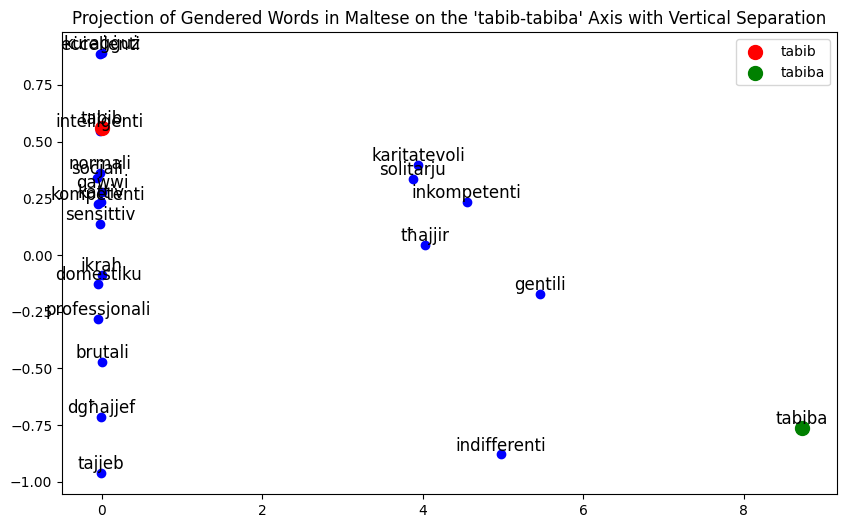}
        \caption{BERTu after debiasing.}
    \end{subfigure}
    \caption{t-SNE visualization of BERTu’s word embeddings for the gendered pair tabib-tabiba (Maltese for ‘doctor’ in male and female forms) before and after applying CDA \cite{lu2018} In the baseline model, tabiba (female doctor) is closer to inkompetenti (incompetent), while tabib (male doctor) is near kompetenti (competent). After debiasing, the expected overlap between tabib and tabiba is not observed—the words remain significantly distant, suggesting that gender distinctions persist in BERTu’s representations. The uneven distribution of adjectives indicates that feminine terms may still be marginalized.}
    \label{fig:bertu_tsne_tabib}
\end{figure*}

\begin{figure*}[h!]
    \centering
    \begin{subfigure}[b]{0.45\textwidth}
        \centering
        \includegraphics[width=0.9\linewidth]{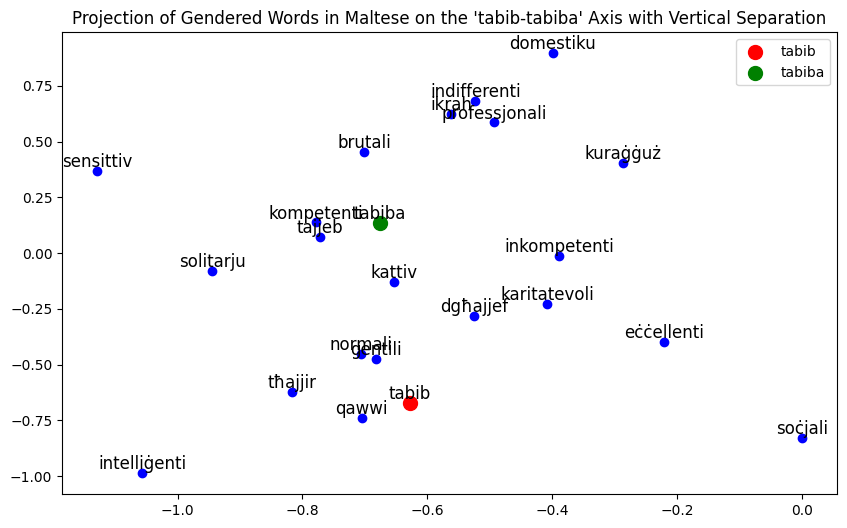}
        \caption{mBERTu t-SNE for `tabib-tabiba'.}
    \end{subfigure}
    \hfill
    \begin{subfigure}[b]{0.45\textwidth}
        \centering
        \includegraphics[width=0.9\linewidth]{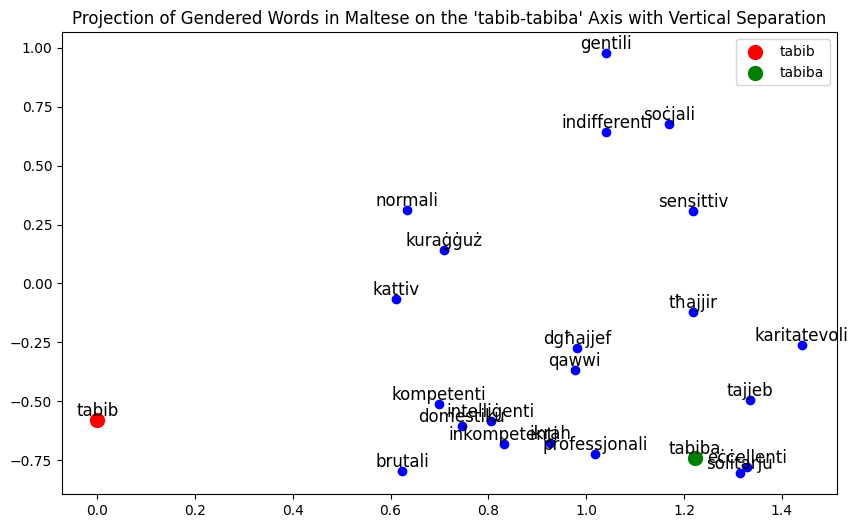}
        \caption{mBERTu after debiasing.}
    \end{subfigure}
    \caption{t-SNE visualization of word embeddings for the gendered pair "\textit{tabib-tabiba}" (Maltese for ‘doctor’ in male and female forms) using mBERTu before and after debiasing using CDA \cite{lu2018}. Compared to BERTu, mBERTu shows a noticeably less biased representation, likely due to its multilingual training. \textit{Kompetenti} (competent) appears closer to \textit{tabiba}, and its antonym is more evenly distributed between the gendered terms. After debiasing, adjectives like \textit{kompetenti}, \textit{professjonali} (professional), and \textit{intelliġenti} (intelligent) are more centered between \textit{tabib} and \textit{tabiba}, indicating reduced bias. However, some adjectives, such as \textit{soċjali} (social) and \textit{sensittiv} (sensitive), remain distant, suggesting that subtle gender associations persist.}
    \label{fig:mbertu_tsne_tabib}
\end{figure*}

\paragraph{Observations}

Both BERTu and mBERTu exhibit gender bias, with monolingual models displaying stronger biases. Occupational bias and societal stereotypes underlie these patterns. CDA proved to be the most effective debiasing method, although grammatical issues arose due to Maltese morphology. Dropout Regularization showed moderate success, primarily benefiting multilingual models. GuiDebias underperformed for Maltese, while Auto-Debias improved monolingual models but sometimes increased explicit bias in multilingual models.

The discrepancies between CrowS and SEAT scores underscore the need for using multiple evaluation metrics, as noted in \cite{woo2023}. Bias mitigation in morphologically rich, low-resource languages like Maltese necessitates tailored approaches that strike a balance between bias reduction and linguistic integrity.

\section{Visual Evaluation}

Inspired by \citeauthor{bolukbasi2016}~\citeyearpar{bolukbasi2016}, we use t-SNE plots to visualise the latent semantic space of gender-triggering adjectives in Maltese. This projection of high-dimensional embeddings helps identify gender bias by analysing how gendered terms cluster. Given that Counterfactual Data Augmentation (CDA) yielded the best debiasing results, we present visualisations for BERTu and mBERTu before and after applying CDA. This was done using three gender word-pairs: "\textit{tabib-tabiba}" (doctor), "\textit{avukat-avukata}" (lawyer) and "\textit{għalliem‐għalliema}" (teacher). The t-SNE plots for BERTu and mBERTU using \textit{tabib-tabiba} can be found in Figures \ref{fig:bertu_tsne_tabib} and \ref{fig:mbertu_tsne_tabib}. The remaining figures are included in Appendix \ref{appendix:tsne}.

The t-SNE visualisations for gendered word pairs in BERTu and mBERTu reveal persistent gender bias in the monolingual model, while the multilingual model exhibits more balanced representations.  
For \textit{tabib-tabiba} (doctor) and \textit{avukat-avukata} (lawyer), baseline BERTu shows clear gendered associations, with \textit{tabiba} and \textit{avukata} (female forms) closely linked to \textit{inkompetenti} (incompetent), while \textit{tabib} and \textit{avukat} (male forms) are associated with \textit{kompetenti} (competent). Additionally, positive and professional adjectives tend to cluster around male terms, reinforcing societal stereotypes. In contrast, baseline mBERTu displays a more diverse distribution, suggesting that multilingual exposure mitigates some of these biases.  

After applying CDA, BERTu still exhibits incomplete debiasing, as \textit{tabib} and \textit{tabiba} remain significantly distant in embedding space, and professional adjectives continue to favour male forms. Similarly, \textit{avukat} retains closer ties to positive adjectives than \textit{avukata}, indicating that bias is reduced but not eliminated. Meanwhile, mBERTu achieves a more neutral distribution post-debiasing, with key adjectives like \textit{kompetenti} and \textit{professjonali} positioned equidistantly between male and female forms, indicating more effective bias mitigation.  

For \textit{għalliem-għalliema} (teacher), baseline BERTu reflects a different stereotype: positive adjectives such as \textit{professjonali} (professional) and \textit{intelliġenti} (intelligent) are more closely linked to \textit{għalliema} (female teacher), while negative terms like \textit{ikrah} (ugly) and \textit{kattiv} (cruel) are associated with \textit{għalliem} (male teacher). This mirrors societal norms that favour women in educational roles while casting men in a harsher light. After CDA, BERTu shows improved gender balance, with \textit{għalliem} and \textit{għalliema} appearing closer together and adjectives more evenly distributed.  

Baseline mBERTu already presents a more neutral representation of \textit{għalliem} and \textit{għalliema}, with positive and negative adjectives distributed more equitably. Post-debiasing, the visualisation remains essentially unchanged, suggesting that mBERTu was less biased to begin with.  



\section{Final observations and Conclusions}

Our analysis revealed that both BERTu and mBERTu exhibit measurable gender bias, with BERTu showing a higher degree of bias. 
This aligns with findings in English models, where monolingual BERT displayed more bias than multilingual mBERT, likely due to the latter’s exposure to diverse linguistic contexts. The bias primarily favoured male-associated terms, particularly in occupational stereotypes, though negative connotations for male terms were also observed, highlighting the complexity of bias patterns.

Among the debiasing techniques tested, CDA was the most effective, significantly reducing bias in both CrowS and SEAT scores. However, it occasionally introduced grammatical errors in Maltese, and the full impact of this technique on the model was difficult to determine without access to appropriate resources. Dropout Regularization had a limited impact, slightly reducing bias in CrowS but increasing implicit bias in BERTu, while showing improvement for mBERTu. GuiDebias did not generalise well to Maltese, increasing bias in both models. Auto-Debias was effective for monolingual models but increased bias in multilingual ones, suggesting its effectiveness depends on the model architecture.

These results highlight the need for multiple evaluation metrics, as different techniques produced conflicting results across CrowS and SEAT. A more nuanced approach is required to fully understand and mitigate bias in language models.

In summary;

\begin{itemize}
    \item \textbf{Counterfactual Data Augmentation (CDA)}: CDA proved to be the most effective debiasing technique for Maltese models among all methods explored in this study, as indicated by the evaluation metrics used.
    
    \item \textbf{Dropout Regularization}: Variations in dropout values resulted in minimal differences in performance. The best results for Maltese were achieved with \( h = 0.2 \) and \( a = 0.15 \) for both monolingual and multilingual models. Dropout Regularization performed considerably better on multilingual models.
    
    \item \textbf{GuiDebias}: This technique did not transfer well to Maltese, and in fact, it increased bias for both models according to our evaluation metrics.
    
    \item \textbf{Auto-Debias}: While Auto-Debias was effective in reducing bias for monolingual models, it increased bias in multilingual models.
\end{itemize}

This research highlights the need for further investigation into bias in multilingual language models, particularly in low-resource languages with complex gender systems, such as Maltese. To aid future work in the area, we publicly share all our experimental and evaluation data, including the Maltese Debiasing Dataset. Future work could significantly expand upon these findings by offering more targeted recommendations, such as identifying which debiasing techniques are better suited for specific tasks (classification versus generation). It would also be beneficial to further analyse what common language features would suit specific debiasing approaches, as well as how debiasing affects LLM performance on NLP downstream tasks such as Named-Entity Recognition and Sentiment Analysis.

While existing debiasing techniques have demonstrated varying levels of effectiveness, our findings underscore the need to refine these methods to better address linguistic and cultural nuances. Future work should focus on developing more robust, language-agnostic debiasing strategies and comprehensive evaluation metrics that can accurately capture different forms of bias across diverse languages.

Additionally, bias research must extend beyond gender and racial biases to include other critical aspects such as age, socioeconomic status, regional dialects, and disability, which remain largely underexplored. Understanding and mitigating these biases is essential for ensuring fairness in AI systems that serve diverse communities.

Our findings contribute to the growing body of research on bias in low-resource languages, emphasising the necessity of adapting mitigation strategies beyond English-centric approaches. As language models continue to shape digital interactions and decision-making systems, it is crucial to prioritise equitable and inclusive AI development. Through continued research and refinement, we can move closer to creating language technologies that are fair, representative, and culturally aware.

\section*{Limitations}

Through this investigation into measuring bias in Maltese LMs and debiasing them using previous debiasing techniques, we acknowledge certain limitations in our work.

\paragraph{CDA} Despite Counterfactual Data Augmentation (CDA) being the best-performing debiasing technique explored for Maltese LMs, the nature of CDA constructs poorly crafted sentences for gendered languages. New sentences are created by pinpointing instances of a word from the wordlist and changing it to the opposite gender, without considering other words, such as verbs, that would need to be modified in a gendered setting to produce a correctly structured sentence. Due to the large number of sentences, it was not feasible to manually correct such sentences, which may hinder the performance of this technique.

\paragraph{Bias Mitigation} Incomplete bias mitigation was seen in the t-SNE visualisations for BERTu. While debiasing reduced certain gendered associations, it did not fully eliminate them. In BERTu, gender distinctions between male and female terms persist even after CDA, suggesting that further refinement is needed. Better results seem to be achieved in mBERTU, the multilingual model.

\paragraph{Debiasing} Impact on Model Utility – Debiasing techniques may unintentionally alter meaningful linguistic relationships, potentially affecting downstream tasks. Evaluating the trade-off between bias reduction and linguistic integrity is crucial. Due to this, we investigated GuiDebias for its attempt at debiasing with minimal effect on the model's language modelling abilities. Still, it was found to transfer poorly for a gendered language such as Maltese.

\paragraph{Dataset and Language Coverage} The debiasing approach was tested on a limited set of gendered word pairs in the Maltese language. Given that biases may vary across different linguistic domains, the findings may not generalise to all contexts or low-resource languages.

\paragraph{Evaluation Constraints} While t-SNE plots provide a useful visual representation of bias, they are inherently subjective. Additional quantitative metrics, such as SEAT or CrowS-Pairs were used to further complement the analysis. It is suggested to use multiple evaluation metrics to form a better understanding of the effects of debiasing on the model. For Maltese, we were limited in metrics, with only CrowS-Pairs being available for Maltese. To aid our investigation, we translated a subset of SEAT files into Maltese; however, future work could aim to expand the selection of metrics.

\paragraph{Multilingual vs. Monolingual Models} The results suggest that multilingual models like mBERTu exhibit reduced bias compared to monolingual models. However, the extent to which multilingual training influences bias remains an open question, requiring further investigation.

\bibliography{custom}

\appendix

\section{Further t-SNE Visualisations}
\label{appendix:tsne}

\begin{figure}[h!]
    \centering
    \begin{subfigure}[b]{0.45\textwidth}
        \centering
        \includegraphics[width=\textwidth]{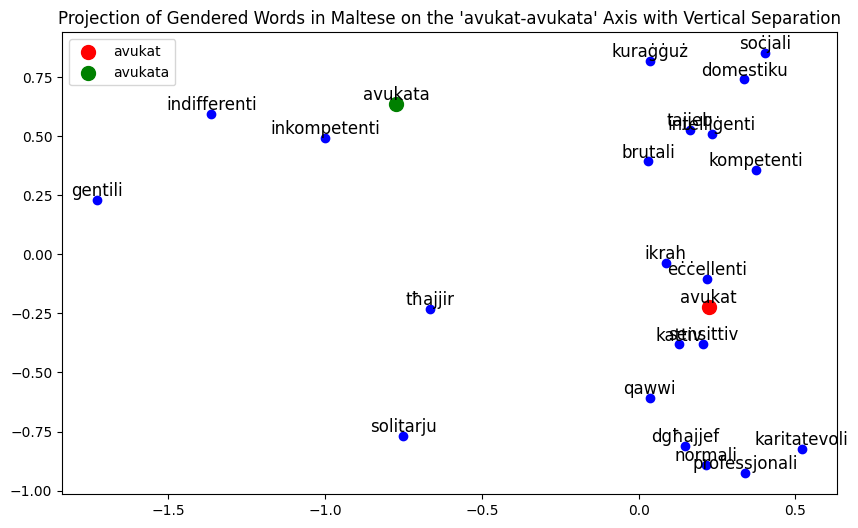}
        \caption{BERTu t-SNE graph for `avukat-avukata'.}
    \end{subfigure}
    \hfill
    \begin{subfigure}[b]{0.45\textwidth}
        \centering
        \includegraphics[width=\textwidth]{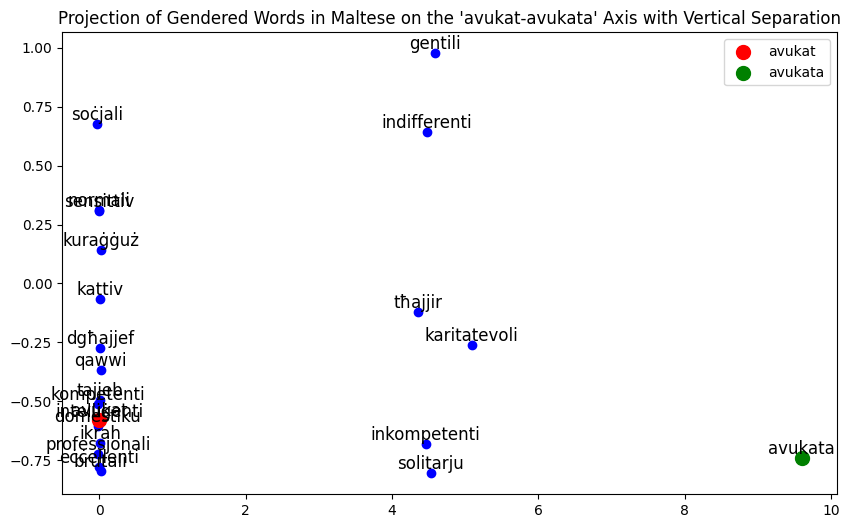}
        \caption{BERTu t-SNE graph for `avukat-avukata' after debiasing.}
    \end{subfigure}
    \caption{t-SNE visualization of BERTu’s embeddings for ‘avukat-avukata’ (lawyer, m-f) before and after CDA.}
    \label{fig:bertu_tsne_avukat}
\end{figure}

\begin{figure}[h!]
    \centering
    \begin{subfigure}[b]{0.45\textwidth}
        \centering
        \includegraphics[width=\textwidth]{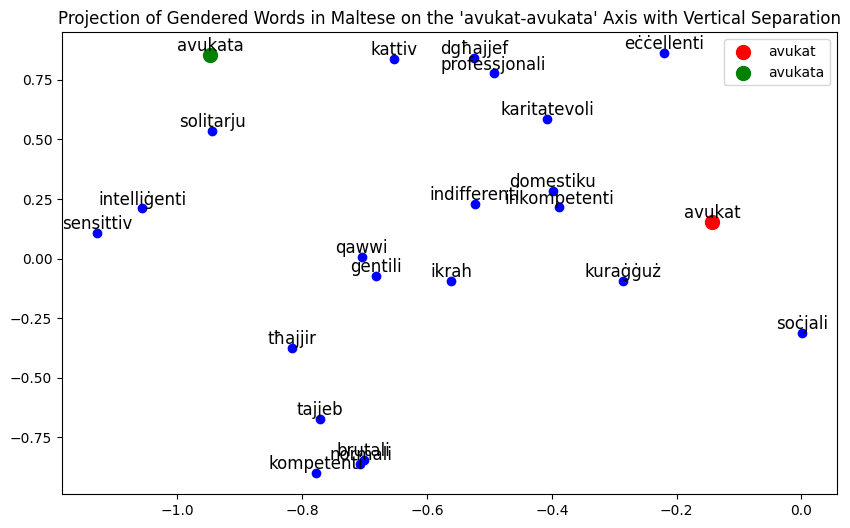}
        \caption{mBERTu t-SNE for `avukat-avukata'.}
    \end{subfigure}
    \hfill
    \begin{subfigure}[b]{0.45\textwidth}
        \centering
        \includegraphics[width=\textwidth]{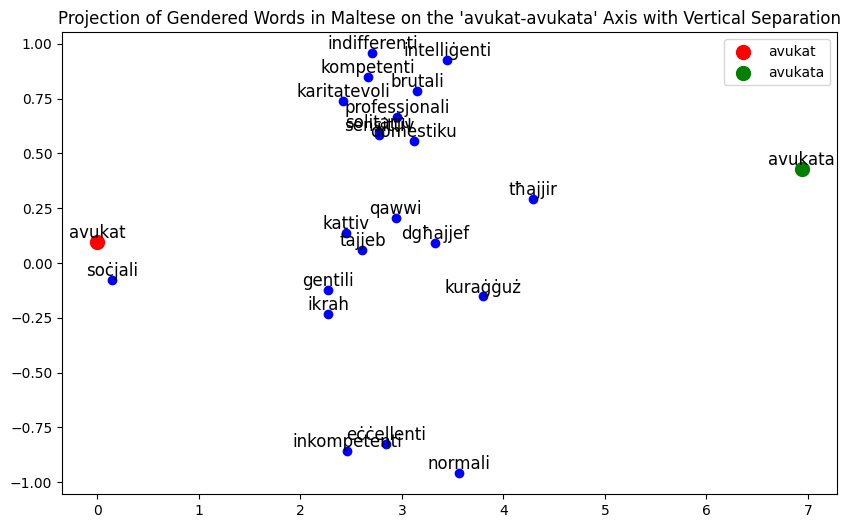}
        \caption{mBERTu t-SNE for `avukat-avukata' after debiasing.}
    \end{subfigure}
    \caption{t-SNE visualization of mBERTu’s embeddings for ‘avukat-avukata’ (lawyer, m-f) before and after CDA.}
    \label{fig:mbertu_tsne_avukat}
\end{figure}

\begin{figure}[h!]
    \centering
    \begin{subfigure}[b]{0.45\textwidth}
        \centering
        \includegraphics[width=\textwidth]{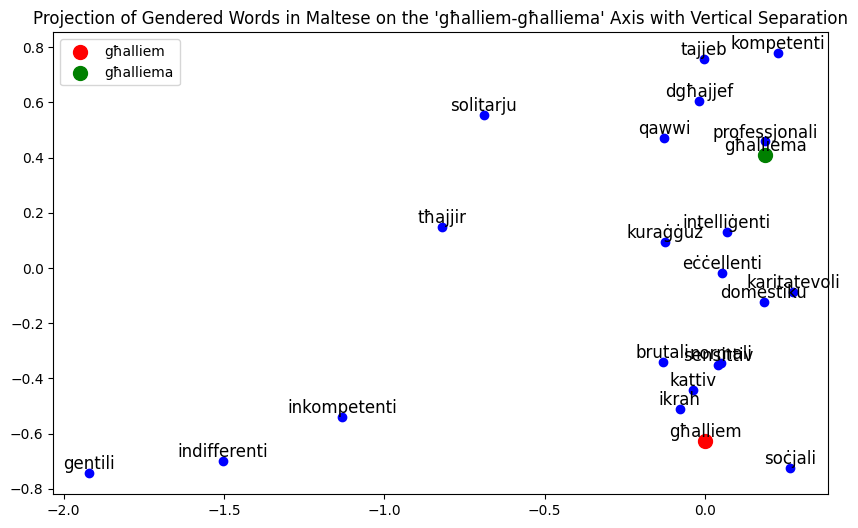}
        \caption{BERTu t-SNE graph for `għalliem-għalliema'.}
    \end{subfigure}
    \hfill
    \begin{subfigure}[b]{0.45\textwidth}
        \centering
        \includegraphics[width=\textwidth]{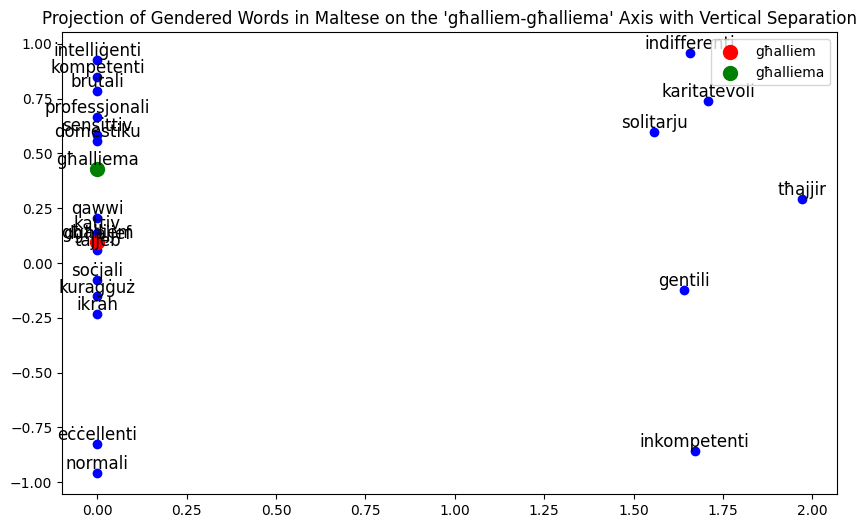}
        \caption{BERTu t-SNE graph for `għalliem-għalliema' after debiasing.}
    \end{subfigure}
    \caption{t-SNE visualization of BERTu’s embeddings for ‘għalliem-għalliema’ (teacher, m-f) before and after CDA.}
    \label{fig:bertu_tsne_ghalliem}
\end{figure}

\begin{figure}[h!]
    \centering
    \begin{subfigure}[b]{0.45\textwidth}
        \centering
        \includegraphics[width=\textwidth]{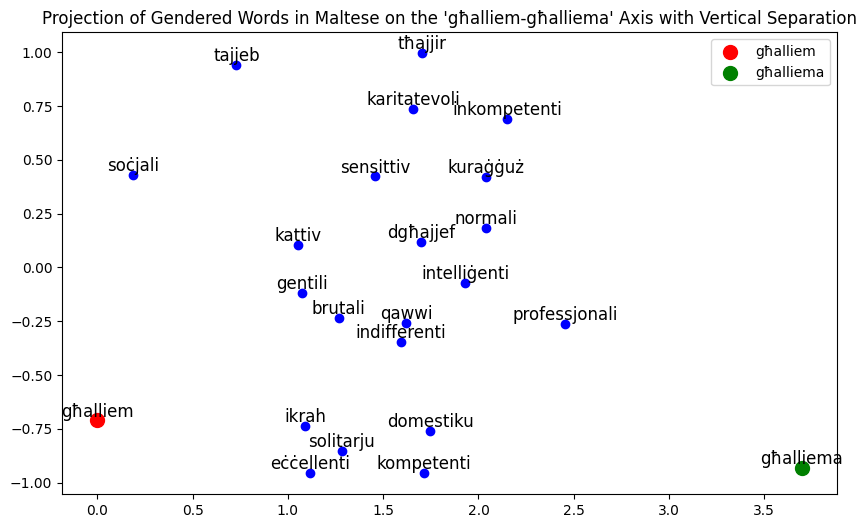}
        \caption{mBERTu t-SNE graph for `għalliem-għalliema'.}
    \end{subfigure}
    \hfill
    \begin{subfigure}[b]{0.45\textwidth}
        \centering
        \includegraphics[width=\textwidth]{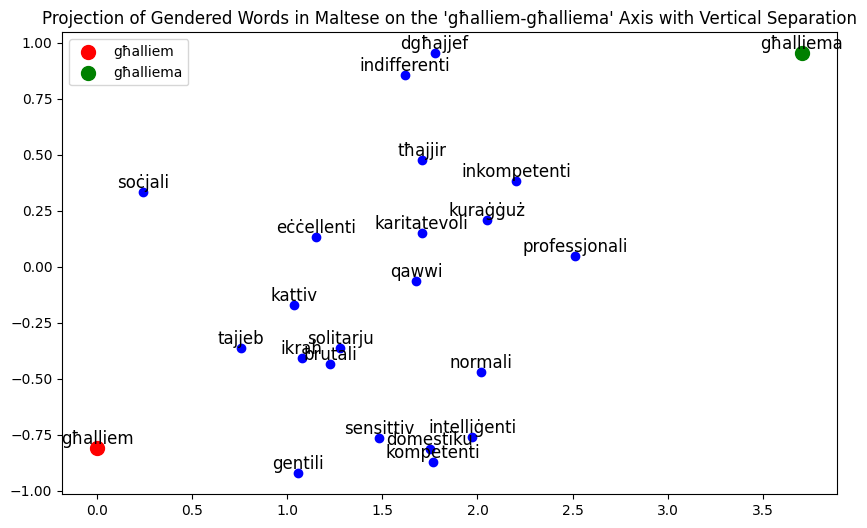}
        \caption{mBERTu t-SNE graph for `għalliem-għalliema' after debiasing.}
    \end{subfigure}
    \caption{t-SNE visualization of mBERTu’s embeddings for ‘għalliem-għalliema’ (teacher, m-f) before and after CDA.}
    \label{fig:mbertu_tsne_ghalliem}
\end{figure}

\end{document}